\PassOptionsToPackage{table}{xcolor}
\documentclass[runningheads]{llncs}

\usepackage[T1]{fontenc}
\usepackage{graphicx}
\usepackage{booktabs}
\usepackage{multirow}
\usepackage{colortbl}
\usepackage{color}
\usepackage[pagebackref,breaklinks,colorlinks]{hyperref}

\usepackage{wrapfig}
\usepackage{orcidlink}
\usepackage{xr-hyper}
\usepackage{algorithm}
\usepackage{algpseudocode}
\usepackage{amsmath,bm}
\usepackage{bbm}
\usepackage{amsfonts}
\usepackage{caption}
\usepackage{subfig}

\usepackage{graphicx}
\usepackage[table]{xcolor} 
 \usepackage{color} 

\definecolor{Gray}{gray}{0.9}
\definecolor{Better}{rgb}{0.18, 0.407, 0.266}
\definecolor{Worse}{rgb}{0.35, 0.35, 0.35}
\definecolor{drakgreen}{rgb}{0.38, 0.67, 0.38}
\definecolor{drakpurple}{rgb}{0.38, 0.27, 0.61}
\definecolor{granate}{rgb}{0.64, 0.16, 0.16}

\newcommand{\our}{\cellcolor{Gray}}

\definecolor{darkolivegreen}{rgb}{0.33, 0.5, 0.18} 

\usepackage[euler]{textgreek}
\usepackage{arydshln}
\usepackage{marvosym}

\usepackage{booktabs,arydshln}

\makeatletter
\def\adl@drawiv#1#2#3{%
        \hskip.5\tabcolsep
        \xleaders#3{#2.5\@tempdimb #1{1}#2.5\@tempdimb}%
                #2\z@ plus1fil minus1fil\relax
        \hskip.5\tabcolsep}
\newcommand{\cdashlinelr}[1]{%
  \noalign{\vskip\aboverulesep
           \global\let\@dashdrawstore\adl@draw
           \global\let\adl@draw\adl@drawiv}
  \cdashline{#1}
  \noalign{\global\let\adl@draw\@dashdrawstore
           \vskip\belowrulesep}}
\makeatother

\begin{document}

\title{Regularized Low-Rank Adaptation for \\ Few-Shot Organ Segmentation}
\titlerunning{Regularized Low-Rank Adaptation for Few-Shot Organ Segmentation}

\author{
Ghassen Baklouti\inst{1,2}\textsuperscript{,\Letter}
\and
Julio Silva-Rodríguez\inst{1}
\and
Jose Dolz\inst{1,2} 
\and \\
Houda Bahig\inst{2}
\and
Ismail {Ben Ayed}\inst{1,2}
}

\authorrunning{G.~Baklouti et al.}

\institute{
\inst{1}ÉTS Montréal \\ 
\Letter {\tt \small \email{ghassen.baklouti.1@ens.etsmtl.ca}} \\
\inst{2}Centre de Recherche du Centre Hospitalier de l’Université de Montréal (CRCHUM) 
}

\index{Baklouti, Ghassen}
\index{Silva-Rodríguez, Julio}
\index{Dolz, Jose}
\index{Bahig, Houda}
\index{Ben Ayed, Ismail}

\maketitle      

\begin{abstract}
Parameter-efficient fine-tuning (PEFT) of pre-trained foundation models is increasingly attracting interest in medical imaging due to its effectiveness and computational efficiency. Among these methods, Low-Rank Adaptation (LoRA) is a notable approach based on the assumption that the adaptation inherently occurs in a low-dimensional subspace. While it has shown good performance, its implementation requires a fixed and unalterable rank, which might be challenging to select given the unique complexities and requirements of each medical imaging downstream task. Inspired by advancements in natural image processing, we introduce a novel approach for medical image segmentation that dynamically adjusts the intrinsic rank during adaptation. Viewing the low-rank representation of the trainable weight matrices as a singular value decomposition, we introduce an $l_1$ sparsity regularizer to the loss function, and tackle it with a proximal optimizer. The regularizer could be viewed as a penalty on the decomposition rank. Hence, its minimization enables to find task-adapted ranks automatically. Our method is evaluated in a realistic few-shot fine-tuning setting, where we compare it first to the standard LoRA and then to several other PEFT methods across two distinguishable tasks: \textbf{base organs} and \textbf{novel organs}. Our extensive experiments demonstrate the significant performance improvements driven by our method, highlighting its efficiency and robustness against suboptimal rank initialization. Our code is publicly available: \href{https://github.com/ghassenbaklouti/ARENA}{https://github.com/ghassenbaklouti/ARENA}.


\end{abstract}

\section{Introduction}
\label{sec:intro}

Despite the remarkable success of deep neural networks for automatic volumetric organ segmentation \cite{litjens2017survey}, these have shown limited flexibility. Concretely, they are usually specialized in specific tasks, requiring large annotated datasets for training and heavy computational demands. As a result, medical image segmentation has been hindered by the necessity of assembling large, curated datasets for deployment \cite{Chen2022}, which is particularly expensive due to the manual labor effort required for annotating volumetric data \cite{wasserthal2023totalsegmentator}. However, a paradigm shift is underway, led by \textit{foundation models}, and more particularly medical-specialized pre-trained models \cite{silva2025foundation,suprem,wang2022medclip,CONCH}. These models are pre-trained on large-scale datasets that leverage heterogeneous, multi-domain samples and tackle multiple tasks within specific medical domains. Thanks to such intensive pre-training, they have demonstrated remarkable generalization properties \cite{radford2021learning,peft_volumetric,suprem}, as well as flexible adaptation to downstream tasks, while requiring minimal supervision \cite{ShakeriMICCAI24,peft_volumetric}. These properties highlight foundation models as a promising path toward a more efficient deployment of automatic segmentation solutions \cite{Malwina2022,Moor2023,foundMed}, with recent open-access models developed for volumetric organ segmentation \cite{UniversalModel,multitalent,peft_volumetric,suprem}, which are pre-trained on an assembly of annotated datasets segmenting multiple base structures. Nevertheless, beyond the significance of providing novel foundation models, exploring how these can be efficiently adapted to novel tasks is paramount. More particularly, \textit{few-shot adaptation} \cite{peft_volumetric}, in which a small number of annotated volumes is required, is of special interest in healthcare, since each institution has limited time, budget, and particular clinical purposes, and the number of available annotated samples is usually limited. However, how to optimally adapt volumetric segmentation foundation models in this setting remains undefined, while being critical for the practical deployment of these models.

\noindent\textbf{\textit{What model parameters to fine-tune?}} This question is essential as the choice might significantly affect the performance. Indeed, in the abundant literature on few-shot classification, most of the existing methods operate on the output embedding space \cite{chen2019closer,boudiaf2020information,HuangCVPR24,ShakeriMICCAI24}, a process often referred to as linear probing. This involves fine-tuning the weights of the last linear-classifier layer while freezing the rest of the network, a common strategy in the context of few-shot segmentation \cite{boudiaf2021few,hajimiri2023strong}. Indeed, full fine-tuning (FFT), i.e., fine-tuning all the learnable parameters of the model, is widely 
avoided in the context of few-shot image classification, as it is prone to over-fitting, substantially degrading the performances \cite{chen2019closer,boudiaf2020information}. Moreover, along with the rise of large foundation models, both in computer vision \cite{radford2021learning} and medical imaging \cite{moor2023foundation,wang2022medclip,silva2025foundation,kirillov2023segment}, which involves millions or even billions of parameters, FFT has becomes less appealing in practice, as it requires substantial computational and memory resources.

\noindent\textbf{\textit{Parameter-Efficient Fine-Tuning (PEFT)}} has emerged as an alternative to address the limitations of FFT. Recently popularized in NLP \cite{he2022towards}, and later adopted in computer vision \cite{gao2024clip,zhang2022tip,zhou2022learning}, PEFT fine-tunes only a small, carefully selected \cite{zaken2022bitfit,layernorm} or added \cite{hu2022lora} set of the trainable parameters, making the adaptation of large models lighter (from computation and memory standpoints). Notable PEFT methods include BitFit \cite{zaken2022bitfit}, which updates only the bias terms of transformer-based models and is commonly used in NLP. Moreover, in the recent literature on vision-language models, PEFT methods have shown excellent performance in few-shot regimes. This includes the seminal work of CoOp, which pioneered prompt learning in VLMs \cite{zhou2022learning}, as well as Adapters \cite{zhang2022tip} and, more recently, Low-Rank Adaptation (LoRA) \cite{zanella2024low}. These recent developments challenged the status quo in few-shot image classification, showing that going beyond linear probing, i.e., updating the inner representations of the models or the input text prompts, could be beneficial. Given the growing success of PEFT in vision and NLP, there is an increasing interest in its application to medical domains and in the few-shot learning settings. This includes, for instance, the recent medical vision-language benchmark in \cite{ShakeriMICCAI24}, as well as the few-shot parameter-efficient fine-tuning framework introduced in \cite{peft_volumetric} for volumetric organ segmentation, and which closely relates to the setting explored in this work.

\begin{figure*}[t!]
    \begin{center}
        \begin{tabular}{lll}

         \includegraphics[width=.48\linewidth]{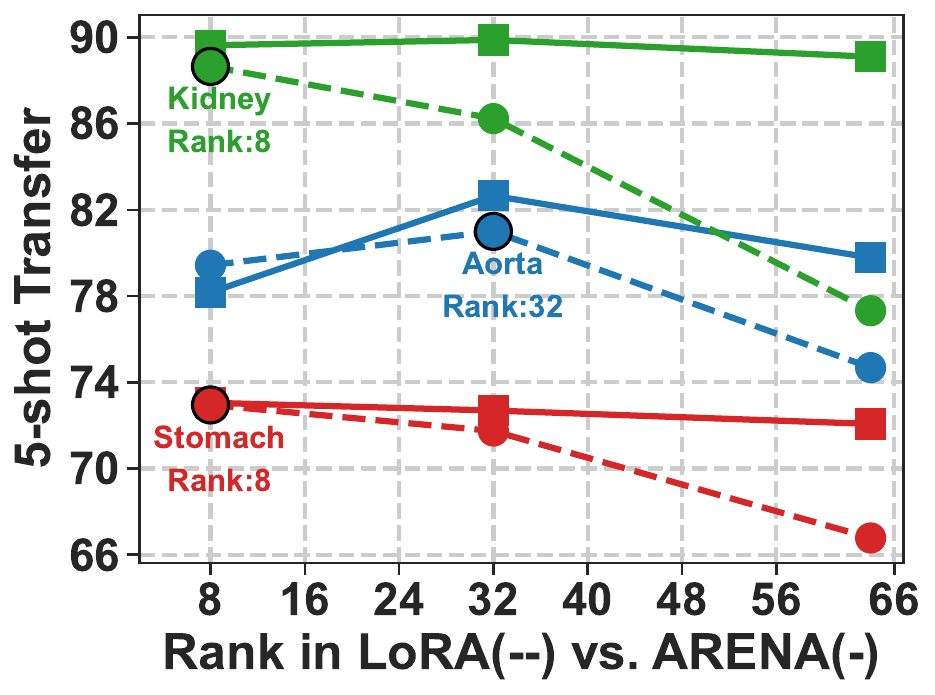} & 
         & \includegraphics[width=.48\linewidth]{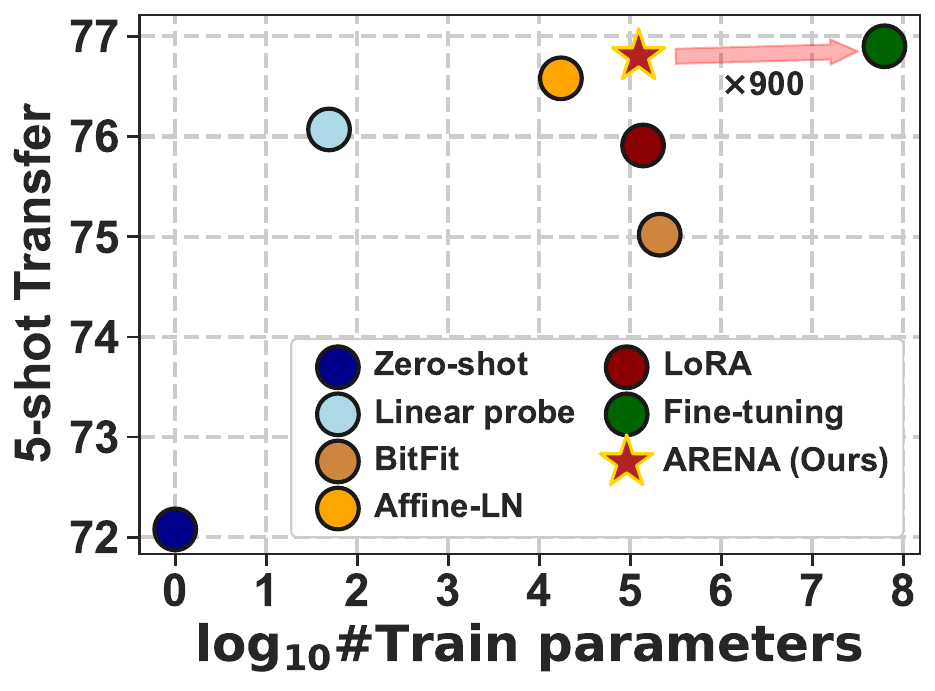} \\

          \multicolumn{1}{c}{\textbf{(a) Robustness}} & & \multicolumn{1}{c}{\textbf{(b) Parameter efficiency}} \\

        \end{tabular}
    \caption{\textbf{Adaptive-LoRA.} We introduce a novel few-shot PEFT technique for \textbf{A}daptive \textbf{R}ank S\textbf{e}gme\textbf{n}t\textbf{a}tion (\textbf{ARENA}) that is (a) \textit{robust to rank initialization} and (b) \textit{enhances the parameter efficiency vs. performance trade-off}.}
        \label{fig:motivation}
    \end{center}
\end{figure*}

\noindent\textbf{\textit{Low-rank adaptation}} is a subcategory of PEFT methods, which integrates additional adaptable low-rank matrices to approximate the weight matrices during adaptation, while keeping the original model weights fixed. This approach was first introduced in the seminal work in \cite{hu2022lora} for NLP tasks, and has since inspired various extensions \cite{valipour2023dylora,zhang2023adalora,ding2023sparse,chavan2023one,dettmers2023qlora}. Following its success in NLP, LoRA has recently garnered increasing attention in computer vision, with the development of several promising approaches. A notable extension is the recent CLIP-LoRA method \cite{zanella2024low}, which customized low-rank adaptation to vision-language models, showing highly competitive performances in few-shot classification. Technically, the standard LoRA baseline approximates the incremental updates $\Delta {W}$ of the pre-trained weights ${W_0}$ as the product of two low-rank matrices, ${ A}$ and ${ B}$. This low-rank modification operation is defined as ${W} = {W_0} + \Delta {W} = {W_0} + {B}{A}$ where ${ A} \in \mathrm{R}^{r\times n}$ and ${B} \in \mathrm{R}^{m \times r}$, with $r$ representing the intrinsic rank, which is typically much smaller than ${m}$ and ${n}$ (i.e., $ r<< ({m},{n})$). Following this framework, only the learnable parameters of matrices $A$ and $B$ are optimized, while the original model parameters remain fixed. Despite its efficiency in reducing the number of trainable parameters and its good empirical performance, LoRA still operates with a fixed rank throughout the optimization process. This constraint limits its flexibility, as the optimal rank selection may vary across different downstream tasks (see Fig. \ref{fig:motivation}(a)). This limitation is exacerbated in few-shot regimes, where relying on validation data for finding the optimal configuration is unrealistic.

\noindent\textbf{\textit{Contributions.}} Building on the above-mentioned momentum in the PEFT and few-shot literature, we investigate low-rank adaptation in volumetric organ segmentation. Our contributions are: \textit{i}) We highlight the limitations of LoRA regarding model selection in few-shot regimes, motivated by the observation in Fig.~\ref{fig:motivation}(a), which shows that the best rank could change significantly from one task (i.e., an organ in this case) to another; \textit{ii}) We propose \textbf{ARENA}, an \textbf{A}daptive \textbf{R}ank S\textbf{e}gme\textbf{n}t\textbf{a}tion method. Viewing the low-rank representation of the trainable weight matrices as a singular value decomposition, we introduce an $l_1$ sparsity regularizer to the loss function, and tackle it with a proximal optimizer. The regularizer could be viewed as a penalty on the decomposition rank. Hence, its minimization enables finding task-adapted ranks automatically; and \textit{iii}) We provide comprehensive experiments showcasing its benefits w.r.t. LoRA in a strict few-shot regime, especially when transferring foundation models to novel segmentation tasks. 

\section{Methods}
\label{sec:methods}

\noindent\textbf{\textit{Preliminaries.}} Building on the few-shot PEFT setting introduced recently in \cite{peft_volumetric}, this study focuses on adapting a large foundation model for volumetric organ segmentation in a realistic clinical scenario, considering the resource limitations of medical institutions. The adaptation process is designed to be both data-efficient (i.e., in a few-shot regime) and computationally lightweight (i.e., following a PEFT paradigm). The adaptation process relies on only a few labeled examples in the target conditions, a.k.a. the {\em support set}, to tackle real clinical settings where annotated medical data is scarce. 
Each few-shot segmentation task involves a support set containing fully labeled volumetric organ samples, denoted as $D_S = {(X_k, Y_k)}^{K}_{k=1}$, where $k$ (typically $K \leq 16$) represents the total number of support samples used for training. It also contains a single query volume $X$ for inference. To mitigate over-fitting in few-shot regimes and further reduce the computational overhead, we focus on PEFT strategies, in which only a tiny subset of the learnable parameters must be fine-tuned, ensuring efficient model adaptation while preserving high segmentation performance.    

\noindent\textbf{\textit{Proposed Adaptive Rank sEgmeNtAtion (ARENA).}} We consider the settings where we fine-tune a pre-trained model by optimizing a supervised-learning loss function ${\cal L} (W)$ using the labeled support set. 
In our experiments, we use the Dice loss, which is widely deployed in medical image segmentation. However, our ARENA framework is applicable to any loss function. To enable automated adjustment of the rank during the adaptation process, let us write the decomposition of the weight matrices in LoRA in the form of a singular value decomposition (SVD): 
\begin{equation}
    {W} = {W_0} + \Delta {W} = {W_0} + B \,\mbox{Diag}(v)\,A,
\end{equation}
where $v$ is an $r$-dimensional vector containing the singular values, and $A$ and $B$ are two low-rank matrices. In SVD decomposition $B\,\mbox{Diag}(v)\,A$, the number of non-zero elements of the vector of diagonal elements $v$, i.e., $\|v\|_0$, determines the rank of the decomposition \cite{ding2023sparse}. Therefore, we can control the rank by imposing a sparsity regularizer on vector $v$. To do so, we add the following $l_1$ regularizer to the loss function, as minimizing the $l_1$ norm promote vector sparsity:
\begin{equation}
\label{regularizaed-loss}
{\cal L} (A, B, v) + \lambda \| v \|_1 
\end{equation}
where ${\cal L} (A, B, v)$ is the loss function that now needs to be optimized over the blocks of learnable parameters $A$, $B$ and $v$. 

\noindent\textbf{\textit{Optimization.}} To optimize regularized loss \eqref{regularizaed-loss} during adaptation, we proceed with a block-coordinate descent process, which alternates two optimization steps: one fixes vector $v$ and optimizes ${\cal L}$ over $(A,B)$ via standard gradient steps, and the other fixes $(A, B)$ and optimizes ${\cal L}$ over $v$ via proximal steps (to account for the non-smooth $l_1$ regularizer).
More precisely, the updates with respect to $A$ and $B$ at iteration $t$ are defined by standard gradient steps: 
\begin{equation} 
A^{(t+1)} = A^{(t)} - \eta \nabla_A {\cal L}(A,B,v), \quad 
B^{(t+1)} = B^{(t)} - \eta \nabla_B {\cal L}(A,B,v), 
\label{eq: standard back-prob}
\end{equation}
where $\eta$ denotes the learning rate, and $ \nabla_A {\cal L}$ and $\nabla_B {\cal L}$ represent the gradients of the loss function with respect to $ A $ and $B$, respectively. Besides, each vector $v$ is updated using a proximal update step \cite{beck2009fast}, enforcing sparsity through an $l_1$ regularization. 
At each iteration $t$, this update is obtained by solving the following problem:  
\begin{equation} 
v^{(t+1)} = \arg\min_{v} \left( \frac{1}{2\eta_t} \| v - (v^{(t)} - \rho \nabla_v {\cal L}(A,B,v) ) \|_2^2 + \lambda \| v \|_1 \right). \label{eq:proximal-optimization}
\end{equation}
where $\nabla_v {\cal L}(A,B,v)$ denotes the gradient with respect to $v$. 
The closed-form solution to this optimization problem yields the update rule for $v$, given by:
\begin{equation}
    v^{(t+1)} = \text{prox}_{\eta_t \lambda \|\cdot\|_1} (v^{(t)}-\rho \nabla_v {\cal L}(A,B,v)) =  \xi(v^{(t)}-\rho \nabla_v {\cal L}(A,B,v), \eta_t \lambda),
    \label{eq: prox step}
\end{equation}
where $\xi(x,\tau)$ is the soft thresholding function defined as:
\begin{equation}
\xi(x, \tau) :=
\begin{cases} 
x - \tau, & x > \tau \\
0, & -\tau \leq x \leq \tau \\
x + \tau, & x < -\tau
\end{cases}
\label{eq:thresholding function}
\end{equation}
This function dynamically regulates the intrinsic rank during adaptation by eliminating small values and scaling down larger ones. It enforces structured sparsity while maintaining the model’s expressive capacity.

\section{Experiments}
\label{sec:exper}

\subsection{Setup}
\label{ssec:setup}

\noindent\textbf{\textit{Foundation model.}} A 3D-SwinUNETR \cite{hatamizadeh2022swin} with open-access weights from a multi-task, multi-dataset supervised pre-training in \cite{peft_volumetric} is employed. Using a supervised learning objective, this model was pre-trained on 2,048 CT scans, addressing the volumetric segmentation of 29 base anatomical structures.

\noindent\textbf{\textit{Adaptation tasks.}} We evaluate the different methods in two types of tasks: \textit{i}) \textbf{Base organs}: anatomical structures used during pre-training, for which the pre-trained model can perform zero-shot predictions, e.g., spleen, left kidney, gallbladder, esophagus, liver, pancreas, stomach, duodenum, and aorta. While we explore binary segmentation in TotalSegmentator dataset \cite{wasserthal2023totalsegmentator}, i.e., one organ is tackled at a time, multi-class adaptation is assessed using FLARE’22 \cite{FLARE22}. \textit{ii}) \textbf{Novel organs}: structures unseen by the network. In particular, we focus on heart parcellation, i.e., individual segmentation of heart-myocardium (MYO), left atrium (LA), right atrium (RA), left ventricle (LV), and right ventricle (RV), from TotalSegmentator volumes. \textbf{Volume pre-processing:} CT scans are standardized following the same pipeline as in pre-training \cite{peft_volumetric}.

\noindent\textbf{\textit{Adaptation training details.}} The model takes as input six patches of size $96\times96\times96$ per volume in each iteration with a batch size of one volume. AdamW is used as optimizer, and a cosine decay learning rate scheduler of 200 epochs is defined for training. The training loss is monitored in the support set, and early stopping is applied upon convergence, assuming
its relative improvement over $20$ epochs does not exceed $1\%$ of the loss value at the beginning of that period.

\noindent\textbf{\textit{PEFT implementation details.}} Following \cite{peft_volumetric}, the decoder is frozen when transferring to known tasks, while it is entirely updated for novel organs. Hence, PEFT methods are incorporated uniquely into the encoder. For the proposed \textbf{ARENA}, the low-rank modifications are applied to the key-value layers of each Transformer block. The matrices $A$ and $B$ follow the standard LoRA initialization, while the gating vector $v$ is initialized from a uniform distribution over the interval $[-1 , 1]$. During optimization, the low-rank matrices are updated using a base learning rate of $10^{-3}$. The gating vector parameters are adjusted to $\lambda = 0.5$ and $\rho = 0$ according to the updating rule in Eq. \ref{eq: prox step}. Across all experiments, the initial rank of ARENA is set to $8$, unless otherwise specified.

\noindent\textbf{\textit{Baselines.}} We include: \textit{i}) black-box strategies, i.e., linear probing; \textit{ii}) full fine-tuning of the pre-trained model (FFT), and \textit{iii}) popular PEFT methods, i.e., selective tuning methods such as BitFit \cite{zaken2022bitfit}, affine layer normalization (Affine-LN) \cite{layernorm}, and additive vanilla LoRA \cite{hu2022lora} and AdaLoRA \cite{zhang2023adalora}. We adhere to the settings provided in \cite{peft_volumetric}, except where explicitly stated, regarding the specific hyper-parameters. \textbf{Evaluation protocol.} The train/test splits in \cite{peft_volumetric} are employed. From training, $K\in\{5, 10\}$ labeled examples are retrieved for training in the few-shot data regime, following the proposed realistic, validation-free adaptation. The test data remains fixed across different trainings, and evaluation is performed through DICE score. Results are averaged across 3 random seeds. 

\subsection{Results}

\begin{table} [t!]
    \setlength{\tabcolsep}{3.5pt}
    \caption{\textbf{Transfer performance to new tasks on TotalSegmentator.} As stated in Section \ref{ssec:setup}, each method is combined with decoder fine-tuning.}
    \label{tab:results_novel_totalseg.}
    \scriptsize
    \centering
    \begin{tabular}{llcccccc} 
    \toprule
         & 
         \multicolumn{1}{c}{Method} & 
         {MYO}& {LA}&{RA}& {LV}&{RV} &{\bf{Avg.}}\\ \midrule
         \multirow{7}{*}{\rotatebox{90}{\textbf{5-shot}}}
         & Linear probe                  & \underline{51.98} &38.99 &40.35 & \underline{53.27} &31.08 &43.13 \\
         & BitFit \cite{zaken2022bitfit} & 51.53 &39.01 &40.19 & \underline{53.27} &31.03 &43.01    \\
         & Affine-LN \cite{layernorm}    & 51.68 & 38.82 & 40.08 & \textbf{53.34} & 31.06 & 43     \\
         & FFT                           & \bf 52.03 &\underline{43.98} &\underline{49.91} &51.22 &33.03 &\underline{46.03}  \\ \cdashlinelr{2-8}
         & LoRA \cite{hu2022lora}        & 41.83 &36.53 &45.67 &43.42 &37.05 &40.90 \\
         & AdaLoRA \cite{zhang2023adalora}        & 50.28 & 43.59 &37.35 &48.53 &\underline{38.73} &43.70 \\
         & \our ARENA (\textit{Ours})    & \our 48.32 & \our \bf 51.97 & \our \bf 54.38 & \our 50.65 & \our \bf 43.69 & \our \bf 49.80  \\
         \midrule
         \multirow{7}{*}{\rotatebox{90}{\textbf{10-shot}}} 
         & Linear probe                  & \underline {64.50} &63.47 &66.86 &69.12 &62.60 &65.31 \\
         & BitFit \cite{zaken2022bitfit} & 64.18 &64.15 &66.35 &\underline{69.79} &62.61 &65.42    \\
         & Affine-LN \cite{layernorm}    & 64.38 & 63.62 & 66.73 & 69.53 & 62.61 & 65.37    \\
         & FFT                           & 59.07 &54.05 &63.06 &64.38 &59.50 &60.01  \\ \cdashlinelr{2-8}
         & LoRA \cite{hu2022lora}        & 60.31 &65.2 &\underline{78.44} &64.05 &\underline{65.29} &\underline{66.66} \\ 
         & AdaLoRA \cite{zhang2023adalora}        & 61.57 & \underline{68.1} &60.81 &59.76 &55.27 &61.10 \\
         & \our ARENA (\textit{Ours})    & \our \bf 75.29 & \our \bf 81.8 & \our \bf 82.93 & \our \bf 74.2 & \our \bf 74.82 & \our \bf 77.81  \\
         \bottomrule
    \end{tabular}
\end{table}

\noindent\textbf{\textit{Transferability to new tasks (Table \ref{tab:results_novel_totalseg.}).}} First, we assess the transfer learning capabilities of the pre-trained foundation model to segment novel structures. In this setting, FFT fails to scale properly with increasing shots ($K=10$) by performing nearly $-5.0$ points below PEFT methods and linear probing. Furthermore, LoRA does not provide consistent gains w.r.t. these baselines across shots. In contrast, the proposed ARENA provides substantial improvements, with performance gains of $+8.9$ and $+11.2$ over LoRA for $K=5$ and $K=10$, respectively. Hence, ARENA is the best transfer learning strategy, allowing flexible feature reuse that benefits robustness when tackling challenging, novel tasks. 

\noindent\textbf{\textit{Transferability to base tasks in TotalSegmentator (Table \ref{tab:results_base_totalseg}).}} First, one may notice that, in contrast to the general belief, FFT is a robust alternative in the low data regime, with consistent improvements over zero-shot predictions. However, as we later discuss, this alternative requires expensive resources for transferring foundation models. PEFT methods are an appealing alternative, which fall close in performance to full fine-tuning. Notably, selective methods such as the baseline Affine-LN are strong baselines that do not introduce additional hyper-parameters nor require explicit control.
In contrast, the popular LoRA depends on fine-grained model selection and fails to improve over PEFT baselines in the proposed validation-free scenario. The proposed ARENA alleviates this issue, resulting in consistent average improvements over vanilla LoRA of $+0.9$ and $+1.6$ for the two explored few-shot data regimes, surpassing linear probing, BitFit, and Affine-LN. It is worth noting that ARENA also outperforms full fine-tuning for $K=10$, which underscores its improved data scalability.

\begin{table} [h!]
    \setlength{\tabcolsep}{3.5pt}
    \caption{\textbf{Transfer performance to base organs on TotalSegmentator.}}
    \label{tab:results_base_totalseg}
    \scriptsize
    \centering
    \begin{tabular}{llcccccccccc} 
    \toprule
         & 
         \multicolumn{1}{c}{Method} &
         {Spl}& {lKid}&{Gall}& {Eso}&{Liv} &{Pan}& {Sto}& {Duo}& {Aor}&{\bf{Avg.}}\\ \midrule
         & Zero-shot &91.34&89.05&77.18&36.73&93.04&78.15&75.86&44.01&63.35&72.08 \\ \midrule
         \multirow{7}{*}{\rotatebox{90}{\textbf{5-shot}}}
         & Linear probe                  & 91.42 &89.51 &\underline{78.15} &45.98 &92.69 &78.31 &\bf{76.64} &62.88 &69.06 &76.07  \\
         & BitFit \cite{zaken2022bitfit} & \bf{92.00} &88.11 &71.11 & \underline{50.00} &92.38 &78.36 &73.00 &56.80 &73.43 &75.02    \\
         & Affine-LN \cite{layernorm}    & 91.43 & \bf 89.95 & 74.95 & \bf 50.65 & 93.04 & \underline{78.81} & 72.41 & \underline{63.05} & 76.71 & 76.78    \\
         & FFT                           & 91.66 &87.22 &73.52 &45.74 &\bf{93.87} &\bf{80.34} &66.24 &\bf{67.31} &\bf{86.18} &\bf{76.90}  \\ \cdashlinelr{2-12}
         & LoRA \cite{hu2022lora}        & 91.48 &88.65 &77.93 &48.02 &92.81 &75.80 &72.95 &56.15 &\underline{79.43} &75.91 \\  & AdaLoRA \cite{zhang2023adalora}        & 91.28 &89.15 &78.11 &43.76 &92.98 &78.35 &\underline{76.17} &58.42 &66.49 &74.96 \\
         
         & \our ARENA (\textit{Ours})    & \our \underline{91.77} &\our \underline{89.63} & \our \bf{79.14} & \our 49.48 & \our \underline{93.09} & \our 78.24 & \our 73.05 & \our 58.59 & \our 78.17 & \our \underline{76.80}  \\
         \midrule
         \multirow{7}{*}{\rotatebox{90}{\textbf{10-shot}}} 
         & Linear probe                  & \underline{91.72} &\bf{89.78} &78.49 &47.01 &92.16 &78.14 &\bf{76.80} &63.63 &69.91 &76.40  \\
         & BitFit \cite{zaken2022bitfit} & 90.85 &87.68 &75.92 &47.92 &91.85 &79.83 &66.35 & \underline{64.10} &77.98&75.83    \\
         & Affine-LN \cite{layernorm}    & 89.22 & 87.88 & 73.48 & \underline{51.11} & 91.29 & 80.05 & 65.99 & 62.42 & 82.42 & 75.98    \\
         & FFT                           & 89.61 &84.79 &76.07 &\bf{56.82} &90.89 &74.87 &60.78 &\bf{71.29} &\bf{91.81} &\underline{77.44}  \\ \cdashlinelr{2-12}
         & LoRA \cite{hu2022lora}        & 89.94 &89.47 &\underline{80.65} &46.11 &\underline{92.94} &\bf{81.18} &66.41 &61.76 &81.66 &76.68 \\ 
         & AdaLoRA \cite{zhang2023adalora}        & 91.28 &89.27 &80.01 &45.72 &92.90 &78.44 &\underline{76.32} &61.81 &68.19 &75.99 \\ 
         & \our ARENA (\textit{Ours})    & \our \bf{92.28} & \our \underline{89.58} & \our \bf{84.49} & \our 50.83 & \our \bf{93.01} & \our \underline{80.27} & \our 68.35 & \our 62.95 & \our \underline{82.46} & \our \bf{78.25}  \\ \bottomrule
    \end{tabular}
\end{table}

\noindent\textbf{\textit{Generalization across datasets.}} Table \ref{tab:results_base_flare} demonstrates the robustness of the improvements of the proposed ARENA w.r.t. vanilla LoRA. Results in FLARE'22 for multi-class segmentation align with the observation from TotalSegmentator, with ARENA providing average gains of nearly $+1.2$ and $+0.5$.

\begin{table} [t!]
    \setlength{\tabcolsep}{2.5pt}
    \caption{\textbf{Transfer performance on an alternative dataset (FLARE'22).}}
    \label{tab:results_base_flare}
    \scriptsize
    \centering
    \begin{tabular}{llcccccccccc} 
    \toprule
         & 
         \multicolumn{1}{c}{Method} &
         {Spl}& {lKid}&{Gall}& {Eso}&{Liv} &{Pan}& {Sto}& {Duo}& {Aor}&{\bf{Avg.}}\\ \midrule
         \multirow{2}{*}{\rotatebox{0}{\textbf{5-shot}}}
         & LoRA \cite{hu2022lora}        & 85.54 &75.68 &54.59 &73.59 &93.98 &82.72 &74.09 &42.59 &91.15 &74.88 \\ 
         & \our ARENA (\textit{Ours})    & \our \bf 88.06 & \our \bf 75.86 & \our \bf 55.71 & \our \bf 75.04 & \our \bf 94.91 & \our \bf 83.61 & \our \bf 76.21 & \our \bf 43.15 & \our \bf 91.52 & \our \bf \our 76.01  \\
         \midrule
         \multirow{2}{*}{\rotatebox{0}{\textbf{10-shot}}}
         & LoRA \cite{hu2022lora}        & 86.37 &\bf 77.05 &55.06 &73.98 &94.05 &83.17 &77.15 &46.23 &91.42 &76.05 \\ 
         & \our ARENA (\textit{Ours})    & \our \bf 87.69 & \our 76.82 & \our \bf 55.4 & \our \bf 75.00 & \our \bf 95.06 & \our \bf 83.59 & \our \bf 77.35 & \our \bf 46.58 & \our \bf 91.6 & \our \bf 76.57  \\ \bottomrule
    \end{tabular}
\end{table}

\noindent\textbf{\textit{Robustness against bad rank initialization.}} We now dig into the limitations of vanilla LoRA to provide satisfactory validation-free results. We focus on rank initialization, as illustrated in Fig.~\ref{fig:motivation}(a), and explore how the transfer performance relies on this critical choice. We perform ablation studies using several rank initializations, i.e., $r\in\{8, 32, 64 \}$, with five shots for adaptation. For example, for the aorta, the used rank ($r=8$) is suboptimal, and transfer would benefit from more flexibility, e.g., using a rank of 32 (+$1.7$). In contrast, our adaptive low-rank adaptation, ARENA, is more robust to bad rank initialization, as shown in Fig.~\ref{fig:motivation}(a). Consequently, the overall transfer results approximate the greedy rank search in LoRA, underscoring its strength and data efficiency, particularly when adapting models in the practical low-shot regime.

\noindent\textbf{\textit{Comparison to other LoRA variants.}} Our results demonstrate that ARENA consistently outperforms AdaLoRA \cite{zhang2023adalora} on both base and novel organs, yielding average gains of approximately $+2$ and $+16$ DICE points, respectively, for $k=10$. While our method, ARENA, shares the same objective of rank adaptation with other SVD-based LoRA variants such as AdaLoRA~\cite{zhang2023adalora} and DyLoRA~\cite{valipour2023dylora}, it differs from them in how this adaptation is achieved. These methods deploy heuristic rules and manual learning schedules~\cite{zhang2023adalora} or rank sampling~\cite{valipour2023dylora}, whereas our method directly integrates rank adaptation into the training objective through L1 regularization, allowing the effective rank to be learned in an entirely data-driven manner, without extensive hyper-parameter tuning.

\noindent\textbf{\textit{Computational efficiency.}} Figure \ref{fig:motivation}(b) illustrates the performance/efficiency trade-off for TotalSegmentator tasks. One can readily notice that vanilla LoRA falls short in performance despite being more parameter-efficient than FFT, training $900\times$ fewer parameters. In contrast, the proposed ARENA approaches FFT while not incurring additional computational overhead.

\section{Conclusions}
\label{sec:conclusion}

This work has addressed the parameter-efficient adaptation of segmentation models in few-shot regimes. Despite PEFT's computational benefits, we observe that popular strategies, such as LoRA, have particular limitations in few-shot adaptation scenarios, which are common in healthcare applications. These point out to the critical role of certain hyper-parameters in additive PEFT strategies, which control model expressiveness, and to the challenges that model selection involves in low-data regimes. The proposed adaptive low-rank strategy, ARENA, alleviates such a burden by promoting sparsity, and provides consistent gains. In this context, we anticipate that adaptive, validation-free techniques will be pivotal in future works in few-shot segmentation.

\begin{credits}
\subsubsection{\ackname} This work was funded by the Natural Sciences and Engineering Research Council of Canada (NSERC) and the Montréal University Hospital Research Center (CRCHUM). We also thank Calcul Quebec and Compute Canada.
\subsubsection{\discintname} The authors have no competing interests to declare that are relevant to the content of this article

\end{credits}

\bibliographystyle{splncs04}
\bibliography{refs}

\end{document}